# On the Utility of Foundation Models for Fast MRI: Vision-Language-Guided Image Reconstruction

Ruimin Feng[a,b], Xingxin He[a,b], Ronald Mercer[b,c], Zachary Stewart[b,c], and Fang Liu[a,b]

[a] Athinoula A. Martinos Center for Biomedical Imaging, Massachusetts General Hospital, Charlestown, Massachusetts, United States

[b] Harvard Medical School, Boston, Massachusetts, United States

[c] Department of Radiology, Massachusetts General Hospital, Charlestown, Massachusetts, United States

# ABSTRACT

**Purpose:** To investigate whether a vision-language foundation model can enhance undersampled MRI reconstruction by providing high-level contextual information beyond conventional priors.

**Methods:** We proposed a semantic distribution-guided reconstruction framework that uses a pre-trained vision-language foundation model to encode both the reconstructed image and auxiliary information into high-level semantic features. A contrastive objective aligns the reconstructed representation with the target semantic distribution, ensuring consistency with high-level perceptual cues. The proposed objective works with various deep learning-based reconstruction methods and can flexibly incorporate semantic priors from multimodal sources. To test the effectiveness of these semantic priors, we evaluated reconstruction results guided by priors derived from either image-only or image-language auxiliary information.

**Results:** Experiments on knee and brain datasets demonstrate that semantic priors from images preserve fine anatomical structures and achieve superior perceptual quality, as reflected in lower LPIPS values, higher Tenengrad scores, and improved scores in the reader study, compared with conventional regularization. The image-language information further expands the semantic distribution and enables high-level control over reconstruction attributes. Across all evaluations, the contrastive objective consistently guided the reconstructed features toward the desired semantic distributions while maintaining data fidelity, demonstrating the effectiveness of the proposed optimization framework.

**Conclusion:** The study highlights that vision-language foundation models can improve undersampled MRI reconstruction through semantic-space optimization.

**Keywords:** vision-language foundation model, semantic prior, contrastive loss, fast MRI

# 1 INTRODUCTION

Foundation models have emerged as a transformative paradigm in artificial intelligence.[1] These models are pre-trained on gigantic datasets using self-supervised learning, where pretext tasks such as masked language modeling,[2] autoregressive next-token prediction,[3] and denoising[4] are employed to capture semantic and contextual structures from raw data. As a result, the learned representations exhibit strong transferability across a wide range of downstream tasks.[5] The rapid progress of foundation models has been largely driven by the transformer architecture,[6] which first demonstrated remarkable success in natural language processing with models such as BERT[2] and ChatGPT[3]. This paradigm was later extended to computer vision through transformer-based designs, including the Vision Transformer[7] and Swin Transformer.[8] Recently, vision-language foundation models align image and textual features through contrastive learning[9], enabling joint understanding across modalities and demonstrating impressive capabilities in tasks such as image understanding and text-to-image generation.[10] These models operate within a shared semantic space that provides a latent representation capturing abstract relationships and organizes information according to perceptual and conceptual similarity in a modality-agnostic manner.[9,11] This shared representation enables seamless knowledge transfer between visual and textual domains, substantially broadening the types of features that can be leveraged in downstream tasks.

Beyond natural images, foundation models are increasingly being applied to medical imaging, where they have shown promising results in tasks such as segmentation,[12,13] disease classification,[14] and automated diagnosis.[15,16] Despite these advances, their applications to solving inverse problems remain relatively underdeveloped, for example, in the reconstruction of fast MRI, which presents unique challenges due to undersampled k-space measurements and the need for strict adherence to imaging physics. These challenges have led current fast MRI reconstruction methods to rely on priors from the inherent imaging physics and other forms of prior knowledge. For example, parallel imaging exploits the information redundancy of multiple receiver coils to recover images from undersampled k-space data.[17,18] Compressed sensing relies on non-uniform k-space undersampling to generate incoherent artifacts, which can be effectively suppressed by imposing priors such as total variation and transform-based sparsity,[19–21] dictionary learning,[22,23] low-rank[24,25] and structured low-rank models.[26,27] Deep learning approaches further introduce powerful data-driven priors that capture statistical image regularities from large-scale datasets,[28] through either supervised[29–36] or self-supervised[37,38] training. More recently, generative diffusion models have emerged to provide priors by learning the distribution of fully sampled images and incorporating physics-based constraints during reconstruction.[39–43] In addition, approaches such as deep image prior[44,45] and implicit neural representation[46,47] utilize the inherent inductive biases of neural networks as implicit regularizers. While these approaches have substantially advanced MRI reconstruction, the priors they employ are generally derived from image-level representations and are shaped by the characteristics of the limited training data. Such priors are effective at capturing low- to mid-level image features, including edges, textures, and spatial smoothness, yet they provide limited access to higher-level semantic information that aligns with human understanding of MRI anatomy and pathology.[48] Moreover, as the priors in task-specific models are learned from datasets of limited scope, their generalizability to imaging domains or acquisition conditions that differ from the training distribution may be reduced.[49]

In this study, we investigate integrating a pre-trained vision-language foundation model to enhance MRI reconstruction. Such models learn shared semantic representations across visual and textual domains. This capability provides access to high-level features of image content and, at the same time, broadens the available prior knowledge by incorporating information from not only existing MR image datasets but also complementary multimodal sources, including other imaging modalities and natural language descriptions. To leverage these advantages, we propose a general optimization framework that operates within the semantic space and aligns the reconstructed representations with the desired target distribution. By fusing these complementary sources of information, our method enriches the contextual knowledge guiding undersampled MRI reconstruction and introduces a preliminary mechanism for incorporating natural-language human input into the reconstruction process. The main contributions of this work are summarized as follows:

1. **Semantic prior integration**: We demonstrate how a large-scale vision-language foundation model, pre-trained on extensive image-text datasets, can enhance MRI reconstruction by incorporating semantic priors into the reconstruction process.

2. **Semantic-space optimization framework:** We propose a general optimization objective formulated in the semantic embedding space, which can be integrated with a variety of deep learning-based reconstruction approaches.

3. **Language-guided reconstruction:** We explore the feasibility of using natural language

as auxiliary guidance to extend the semantic representations for MRI reconstruction. Compared with purely image-based priors, textual descriptions offer greater accessibility and flexibility, thereby enriching the prior information available for undersampled MRI reconstruction.

## 2 THEORY AND METHODS

### 2.1 Overview of Vision-language Foundation Models

**Figure 1(A)** illustrates a general overview of vision-language foundation models for image understanding. The input consists of an image paired with a language instruction. The image is processed by an image encoder, often implemented with a Vision Transformer architecture.[50] As shown in the yellow box in **Figure 1(A)**, the image encoder divides the input image into non-overlapping patches, flattens each patch, and maps them through a learned linear projection. Positional information is then added to these patch embeddings, which are subsequently passed through a stack of transformer layers composed of multi-head self-attention and multilayer perceptrons (MLPs). This structure enables the extraction of both fine-grained local details and long-range contextual relationships. The accompanying language instruction is processed by a text tokenizer, which converts the sentence into a sequence of subword tokens. Each token is mapped to a unique index within a predefined vocabulary and then transformed into a continuous embedding space. The visual and textual embeddings are subsequently aligned or fused within a shared semantic space through mechanisms such as multimodal transformers, cross-attention modules, or contrastive objectives, depending on the specific model design. When required to generate a textual response, the resulting output token sequence is converted back into natural language by a text de-tokenizer that maps predicted token indices to their corresponding words.

In addition to these architectural components, most vision-language foundation models incorporate contrastive learning objectives during pretraining, which help the model learn meaningful and well-organized representations from large-scale multimodal data. In contrastive learning, semantically related samples are treated as positive pairs, whereas unrelated samples serve as negative pairs. A commonly used contrastive objective is the InfoNCE loss,[51] which encourages embeddings of positive pairs to be close while pushing apart those of negative pairs. Given $N$ samples in a batch, the InfoNCE contrastive loss is defined as:

$$\mathcal{L}_{con} = -\frac{1}{N}\sum_{i=1}^{N}\log\frac{\exp\left(\text{sim}(e_i, e_i^+)/\tau\right)}{\sum_{j=1, j\neq i}^{N}\exp\left(\text{sim}(e_i, e_j^-)/\tau\right)}, \qquad (1)$$

where $e_i$ represents the embedding of the $i$th sample, $e_i^+$ and $e_j^-$ denote positive and negative embeddings relative to $e_i$. $\tau$ is the temperature parameter that controls how strongly the model distinguishes between positive and negative pairs. A small $\tau$ makes the similarities more "sharp," meaning that even small differences between embeddings are greatly amplified. The similarity term $\text{sim}(\cdot)$ typically refers to the cosine similarity, defined as:

$$\text{sim}(a, b) = \frac{a \cdot b}{\|a\|\|b\|} \qquad (2)$$

which measures the angular closeness between two vectors independent of their magnitudes. While the original InfoNCE formulation restricts each sample to having a single positive counterpart, later variants generalize it to multi-positive settings,[52,53] in which several semantically similar samples are simultaneously treated as positives:

$$\mathcal{L}_{con}(e_i, \mathcal{P}, \mathcal{N}) = -\frac{1}{N}\sum_{i=1}^{N}\log\frac{\sum_{p\in\mathcal{P}}\exp\left(\text{sim}(e_i, p)/\tau\right)}{\sum_{p\in\mathcal{P}}\exp\left(\text{sim}(e_i, p)/\tau\right) + \sum_{n\in\mathcal{N}}\exp\left(\text{sim}(e_i, n)/\tau\right)}, \qquad (3)$$

where $\mathcal{P}$ represents the set of positive embeddings, $\mathcal{N}$ is the set of negative embeddings. Through minimizing this loss, the model organizes the multimodal embedding space such that samples with similar semantic content lie closer together.

To obtain such structured semantic representations, vision-language foundation models are usually pretrained on large-scale multimodal datasets that include diverse images and textual descriptions. A typical example is the Janus model[54] (https://github.com/deepseek-ai/Janus), which is one of the most recent open-source foundation models. Janus uses a multimodal transformer-based large language model to process visual and textual embeddings together and is pretrained on about 1.25 million image-text pairs, allowing for both perceptual and contextual semantic understanding. This Janus model serves as the basic foundation model in this study.

## 2.2 Leveraging Vision-language Foundation Models for MRI Reconstruction

The pre-trained vision-language foundation model introduces unified semantic representations that capture high-level visual structures and linguistic context. These rich semantic features offer complementary prior knowledge to assist undersampled MRI reconstruction. To incorporate such priors into a reconstruction framework, we formulate the problem using the conventional MRI forward model. Let:

$$S_j = \mathbf{MFC}_j I + \delta_j, \quad (4)$$

where $S_j$ is the undersampled k-space data of the $j$th coil. $I$ is the vectorized MR image to be reconstructed and $\delta_j$ represents the noise. The matrix $\mathbf{C}_j$ corresponds to the diagonalized coil sensitivity map for the $j$th coil, $\mathbf{F}$ denotes the Fourier transform, and $\mathbf{M}$ is the diagonalized sampling mask. Recovering $I$ from $S_j$ can be formulated as solving the problem:

$$\arg\min_I \frac{1}{2}\sum_{j=1}^{c} \left\| S_j - \mathbf{MFC}_j I \right\|_2^2 + \lambda \mathcal{R}(I), \quad (5)$$

where $c$ is the total number of receiver coils and the first term enforces data consistency with the acquired measurements. $\mathcal{R}(\cdot)$ denotes the regularization term that incorporates prior knowledge about the reconstructed image $I$, while $\lambda$ controls the contributions between these two components. In this study, we introduce a semantic consistency prior by formulating $\mathcal{R}(\cdot)$ with a contrastive learning objective. The idea is to ensure the semantic embedding of the reconstructed image stays close to semantically relevant examples and far from irrelevant ones within the foundation model's embedding space. This results in the following regularized reconstruction problem:

$$\arg\min_I \frac{1}{2}\sum_{j=1}^{c} \left\| S_j - \mathbf{MFC}_j I \right\|_2^2 + \lambda \mathcal{L}_{con}(e_I, \mathcal{P}, \mathcal{N}), \quad (6)$$

where $e_I$ is the semantic embedding extracted from the current reconstruction $I$. $\mathcal{P}$ and $\mathcal{N}$ denote the sets of positive and negative embeddings, respectively.

Equation (6) establishes a general goal for optimizing semantic space. By leveraging the high-level perceptual and contextual representations provided by the foundation model, this goal guides the reconstruction toward images that better reflect realistic MRI content. Furthermore, the formulation is adaptable and can be incorporated into a range of deep learning architectures for MRI reconstruction. For a given network $f_\theta$ with parameters $\theta$ and input $d$, the overall loss function is formulated as:

$$\mathcal{L}(\theta) = \frac{1}{2}\sum_{j=1}^{c} \left\| S_j - \mathbf{MFC}_j f_\theta(d) \right\|_2^2 + \lambda \mathcal{L}_{con}\left[e_{f_\theta(d)}, \mathcal{P}, \mathcal{N}\right], \quad (7)$$

The network parameters $\theta$ are optimized via gradient backpropagation to minimize $\mathcal{L}(\theta)$, ensuring that the reconstructed images satisfy both the MR data consistency and the high-level semantic consistency imposed by the foundation model.

As shown in **Figure 1(B)**, we investigated three deep learning architectures in this study:

(1) End-to-end mapping[29] with a U-Net: This approach learns a direct transformation from undersampled inputs to reconstructed images.

(2) Unrolled iterative network:[33,37] This method unrolls a variable-splitting iterative reconstruction algorithm into a sequence of learnable stages. Each stage alternates between a neural network update block (e.g., U-Net) and an explicit data-consistency step solved using conjugate gradient descent.

(3) Implicit neural representation (INR) network:[46] This method represents the MR image as a continuous function of spatial coordinates. The function is parameterized by an MLP, which takes spatial coordinates as input and predicts the corresponding image intensities.

All three reconstruction backbones are optimized in a self-supervised manner using the loss function defined in Eq. (7).

## 2.3 Semantic Embedding Extraction from the Foundation Model

**Figure 1(C)** illustrates the detailed process used to extract semantic embeddings from the Janus foundation model for semantic space optimization. Specifically, we consider two scenarios that incorporate different forms of auxiliary information:

(1) Image-only prior: A set of available fully sampled and undersampled MR images is used as auxiliary information to construct the positive and negative prior embeddings. These embeddings are obtained through the image-encoder component $\mathbf{\Phi}_{img}$ of the Janus model, i.e.:

$$\mathcal{P} = \{\mathbf{\Phi}_{img}(I_i^{fully})\}_{i=1}^{M}, \mathcal{N} = \{\mathbf{\Phi}_{img}(I_i^{under})\}_{i=1}^{M} \tag{8}$$

where $M$ is the total number of auxiliary image examples. The reconstructed image $I$ is encoded using the same procedure so that its semantic embedding lies in the same space as the prior embeddings:

$$e_I = \mathbf{\Phi}_{img}(I) \tag{9}$$

(2) Image-language prior: A small set of auxiliary positive and negative MR images (e.g., fully sampled and undersampled) is paired with a language instruction that prompts the model to perform an image-quality assessment, such as *"Determine whether this image is high-quality or low-quality."* As illustrated in **Figure 1(C)**, each auxiliary image is encoded by the image encoder $\mathbf{\Phi}_{img}$, while the accompanying language instruction is converted into token embeddings through the text tokenizer $\mathbf{\Phi}_{txt}$. The visual and textual embeddings are then jointly processed by the large language model $\boldsymbol{\Phi}_{llm}$, which produces a language semantic embedding representing the model's quality-assessment response for the given image (e.g., "high-quality" or "low-quality"). In this scenario, the number of auxiliary MR images can be substantially reduced, potentially leading to an incomplete representation of the underlying semantic distribution. To mitigate this limitation, we expand the semantic space by introducing Gaussian perturbations to the text-derived embeddings. The standard deviation of the perturbation is carefully tuned to broaden semantic coverage while preserving the core semantic identity of the embeddings. This preservation criterion is validated by verifying that the large language model yields responses with equivalent meaning when conditioned on the perturbed embeddings. The process can be formulated as:

$$\mathcal{P} = \{\boldsymbol{\Phi}_{llm}(\boldsymbol{\Phi}_{img}(I_i^{pos}), \boldsymbol{\Phi}_{txt}(T) + \epsilon_k) | i = 1,2,\dots,M, \ k = 1,2,\dots K\},$$
$$\mathcal{N} = \{\boldsymbol{\Phi}_{llm}(\boldsymbol{\Phi}_{img}(I_i^{neg}), \boldsymbol{\Phi}_{txt}(T) + \epsilon_k) | i = 1,2,\dots,M, \ k = 1,2,\dots K\} \quad (10)$$

where $I_i^{pos}$ and $I_i^{neg}$ denote positive and negative image examples, respectively. $T$ represents the language instruction. $K$ is the number of perturbations, and $\epsilon_k \in \text{Gaussian}(0, \sigma)$ denotes the $k$th perturbation. The reconstructed image $I$ is passed through the same image-language encoding pipeline, ensuring that it is mapped into the same language semantic space defined by the model's quality-assessment responses:

$$e_I = \boldsymbol{\Phi}_{llm}\left(\boldsymbol{\Phi}_{img}(I), \boldsymbol{\Phi}_{txt}(T)\right) \quad (11)$$

By extracting high-level semantic priors from the auxiliary information and optimizing the reconstructed embeddings accordingly, the reconstruction process is guided toward the semantic distribution defined by the positive examples. In the image-only setting, this corresponds to aligning the reconstruction with the visual features and anatomical details represented in the fully sampled reference images. In the image-language setting, the reconstruction is steered toward the semantic attributes conveyed by the positive textual descriptions, thereby enabling the model to recover image features consistent with the intended linguistic cues.

## 2.4  Implementation details

In Janus, the vision transformer within the image encoder divides each input image into non-overlapping $24 \times 24$ patches, which are then linearly projected and processed through 24 transformer layers. Each transformer layer maps the input into 1024-dimensional visual embeddings. The large language model also consists of 24 transformer layers and outputs 2048-dimensional embeddings. These embeddings constitute the semantic space within which the reconstruction is optimized. The weights of Janus model are fixed in this study.

For deep learning backbones used in MRI reconstruction, the U-Net architecture, whether used in end-to-end mapping or in an unrolled model, follows a standard encoder-decoder design. Each U-Net contains five resolution levels with two 3×3 convolutional layers per level, followed by batch normalization and ReLU activations. The encoder employs 64, 128, 256, 512, and 1024 feature channels across the five levels, with symmetric decoding pathways and skip connections. The U-Net operates on complex-valued images represented as two input channels corresponding to the real and imaginary components. The output is also produced in two channels representing the reconstructed real and imaginary parts. No activation function is applied to the final output layer. The unrolled iterative model consists of four unrolled stages, and the network parameters are not shared across stages. The MLP in INR consists of eight fully connected layers, each containing 256 hidden units with sine activation functions[55] to improve the network's ability to learn high-frequency information. The final output layer does not apply an activation function. The MLP maps each spatial coordinate to a two-channel output representing the real and imaginary components of the reconstructed image. All inputs are normalized to the range $[-1,1]$ in coordinate space to facilitate stable training.

For the image-only prior, three hierarchical embedding levels extracted from the 1st, 13th, and 24th transformer layers of the image encoder were used to capture low-, mid- and high-level semantic information. The corresponding weights for these three levels were set to 0.005, 0.5, and 1, respectively. For the image-language prior, the large language model's final output was used as the semantic representation, as it provides the most comprehensive and semantically complete information for highly abstract language descriptions. To augment the language semantic distributions, 100 Gaussian noise perturbations ($K=100$) with a standard deviation of $\sigma=0.03$ were added to the textual embeddings. The temperature parameter in the contrastive loss was fixed at 0.07 and kept consistent across all experiments.

All methods were implemented in Python 3.9 using PyTorch 2.8.0 on a workstation equipped with an INTEL(R) XEON(R) PLATINUM 8562Y+ CPU and an NVIDIA H100 GPU with 96 GB of memory. The code is available at: https://github.com/I3Tlab/Foundation-Model-MRI-Reconstruction.

## 2.5 Experimental Settings
### 2.5.1 Evaluation of Image-only Prior

This experiment evaluates the performance of the image-only prior. We used knee (proton density weighted 2D-FSE with fat suppression) and brain (T2 FLAIR) datasets from the FastMRI repository.[56,57] To construct the image embedding distributions for $\mathcal{P}$ and $\mathcal{N}$, a subset of the training data, including 20 subjects for knee and 50 subjects for brain, was used as auxiliary images, while reconstruction was conducted on the corresponding validation datasets (99 subjects for knee data, 69 subjects for brain data) by retrospectively undersampling k-space data. For both the end-to-end U-Net and the unrolled model, the networks were first warm-started on auxiliary images to learn general structural representations and subsequently fine-tuned on the specific undersampled data by minimizing the loss function in Eq. (7), with a low-rank adaptation scheme for fast network convergence.[58] The performance of the proposed semantic priors was compared with the conventional method using total variation (TV) regularization under the same initialization and training settings.

Reconstruction results were assessed using peak signal-to-noise ratio (PSNR), structural similarity index measure (SSIM),[59] Tenengrad,[60] and learned perceptual image patch similarity (LPIPS).[61] Tenengrad measures image sharpness by computing the gradient magnitude, while LPIPS evaluates perceptual similarity based on deep neural network features. To further assess image quality beyond these quantitative metrics, a reader study was conducted in which two experienced musculoskeletal radiologists, both specialized in knee MRI, independently and blindly evaluated the reconstructions from all three methods under two regularization conditions on 20 subjects. Images were randomly presented, and each was scored on a 5-point Likert scale[62] (0=very poor, 5=excellent; higher scores indicate better quality) across four criteria: presence of artifacts, sharpness of anatomical structures, diagnostic confidence, and overall image quality. Artifacts arising from imperfect fat suppression were not considered during scoring, as they originate from the acquisition rather than the reconstruction methods. The reconstruction results were also analyzed in the semantic space by visualizing the embedding distributions using the uniform manifold approximation and projection (UMAP) method.[63]

**2.5.2 Robustness and Generalization of Image-only Prior**

We further evaluated the generalization and robustness of the image-only prior across cases with substantial structural abnormalities or different imaging contrasts, which may not be captured in the auxiliary images. Specifically, four representative scenarios were selected to assess the performance under large anatomical deviations: (1) knee joint replacement, characterized by metal-induced signal voids and altered anatomy; (2) edema, presenting diffuse signal changes and blurred tissue boundaries; (3) ventricular enlargement, reflecting large-scale anatomical deformation; and (4) white matter hyperintensities with tumors, combining both lesion burden and contrast variation. To evaluate robustness across contrast domains, FastMRI knee data using 2D-FSE without fat suppression were also reconstructed using semantic embeddings extracted from 2D-FSE with fat suppression.

**2.5.3 Effectiveness of Image-language Prior**

In this experiment, we investigated the role of language in expanding the distribution of semantic embeddings and its impact on image reconstruction. We used only 3 subjects from the FastMRI knee dataset to construct two types of language-augmented semantic priors:

(i) "High-quality" vs. "Low-quality" semantic space: Embeddings were obtained from fully sampled and undersampled images paired with the instruction "*Determine whether this image is high-quality or low-quality.*"

(ii) "Blurred" vs. "Aliased" semantic space: Embeddings were derived from TV-regularized compressed sensing reconstruction results (causing blurring) and undersampled images paired with the instruction "*Determine whether this image is blurred or aliased.*"

We monitored the reconstruction outputs across iterations, along with the corresponding quality assessment responses generated by the large language model, to illustrate the influence of language embeddings on the reconstruction trajectory. To evaluate the robustness and flexibility of the language instructions, we generated 20 semantically equivalent prompts with varied phrasings, as summarized in **Supporting Information Table S1**. The variability in the resulting reconstructions across these differently worded prompts was then analyzed to assess the consistency of the model's language-conditioned behavior.

## 3 RESULTS

### 3.1 Reconstruction Performance with Image-only Prior

**Figure 2** presents the reconstruction results on knee data using the conventional TV regularization and the proposed semantic prior. At an acceleration rate of R=4 (**Figure 2(A)**), both approaches effectively suppress aliasing artifacts. The zoomed-in views show that the proposed prior better preserves realistic textures in the bone region, producing results that visually resemble the fully sampled reference, whereas TV regularization tends to over-smooth the structures. **Figure**

**2(B)** presents reconstruction results at an acceleration rate of R=6. The displayed image corresponds to a representative knee slice containing a meniscus tear, indicated by the yellow arrow. This pathological feature is depicted with greater clarity in the reconstructions guided by semantic priors. Moreover, the layered cartilage (blue arrow) and the surrounding tissue (green arrow) exhibit improved sharpness and structural definition across all three deep learning methods when incorporating semantic priors. **Figure 3** presents reconstruction results on brain data. At R=4 (**Figure 3(A)**), incorporating semantic priors yields higher visual fidelity and better preservation of fine anatomical details across all three deep learning methods, as shown by the boundaries of the globus pallidus (red arrow). **Figure 3(B)** highlights a lesion region, where the surrounding anatomical structures are delineated more distinctly in the results of end-to-end and unrolled methods guided by semantic priors, as indicated by the yellow arrow.

**Table 1** compares quantitative results for both the knee and brain datasets using the conventional and proposed priors. For the knee data at R=4 and R=6, the end-to-end and unrolled methods with proposed semantic priors achieve better LPIPS and Tenengrad scores than the TV-regularized reconstruction, although they yield lower PSNR and SSIM. The INR model combined with semantic priors demonstrates improved performance across all four metrics. Similar trends are observed on the brain data, integrating semantic priors consistently improves LPIPS and Tenengrad, while producing lower PSNR for all three architectures. For SSIM, the semantic prior produces slightly lower or comparable values in the end-to-end and unrolled methods, and higher values in the INR network. **Figure 4** shows the reader study result comparing the proposed semantic-prior reconstruction with the conventional TV-regularized baseline across four evaluation criteria. The proposed semantic prior yields higher mean scores across all criteria for each network architecture. Statistical testing using the Wilcoxon signed-rank test indicates significant improvements ($p<0.05$) for several comparisons, particularly for the U-Net and unrolled reconstructions. The INR results follow similar trends, although differences did not reach statistical significance.

**Figure 5** visualizes the embedding distributions of undersampled and fully sampled auxiliary images, as well as the reconstructed images across three hierarchical levels in the semantic space. **Figure 5(A)** shows the clustering of data points. The reconstructed embeddings (red) overlap with the fully sampled cluster (orange) across all three levels, while remaining distinct from the undersampled embeddings (blue). **Figure 5(B)** shows the optimization trajectory of a representative example over the course of reconstruction. The trajectories are illustrated using a color gradient, with the initial point shown in blue and the final point marked by a red star. The zoomed-in panels highlight the local optimization paths. These trajectories reveal a consistent convergence from the undersampled distributions toward the fully sampled cluster across all three levels, indicating that the contrastive loss successfully drives the reconstructed representations toward the positive distribution while repelling them from the negative distribution, thereby enforcing semantic consistency with the fully sampled data

## 3.2 Generalization and Robustness of Image-only Prior

**Supporting Information Figure S1** presents the reconstruction results of the three methods, guided by semantic priors, for cases with substantial anatomical abnormalities at a 4× acceleration. Although the auxiliary images used to construct the priors do not include these atypical structures, all three methods are able to recover the major pathological features, including the regions affected by joint replacement, edema, choroid plexus structures within enlarged ventricles, and tumors. In contrast, the zero-filled images exhibit pronounced aliasing and substantial loss of anatomical detail. **Figure 6** shows the reconstruction results for knee data acquired with 2D-FSE without fat suppression, using either conventional regularization or the proposed semantic prior extracted from auxiliary 2D-FSE images with fat suppression. In the zoomed-in region containing a meniscal tear (yellow arrow), the pathological structure and adjacent tissues (red arrow) are more distinctly visualized in the reconstructions guided by the proposed prior. In contrast, conventional regularization tends to produce smoother appearances in these areas, reducing the visibility of fine structural details.

## 3.3 Effectiveness of Image-language Prior

**Figure 7** illustrates the feasibility and effectiveness of the proposed image-language prior in guiding MRI reconstruction at an acceleration rate of 4. **Figure 7(A)** visualizes the embeddings in the language semantic space. The red and black points correspond to high-quality and low-quality embeddings obtained from representative image examples paired with the instruction "*Determine whether this image is high-quality or low-quality.*" The orange and blue points denote the augmented embeddings generated by adding Gaussian perturbations to the original language embeddings of the instruction, demonstrating that this strategy effectively broadens the semantic distribution when only a few image examples are available. The color-gradient trajectory depicts the optimization path of a representative reconstruction, showing a gradual transition of its embedding from the low-quality region toward the high-quality region. **Figure 7(B)** shows the corresponding evolution of the reconstructed images across iterations. The visual quality improves progressively from low-quality intermediate results to a high-quality final reconstruction. The large language model's responses follow the same trend, shifting from "*Low-quality*" at early iterations to "*High-quality*" in later stages.

**Figure 8** illustrates the effectiveness of the proposed image-language prior in a second case, using the "blurred" versus "aliased" semantic setting. Similarly, the prior distributions are generated from representative image examples paired with the instruction "*Determine whether this image is blurred or aliased.*" and are further augmented using Gaussian perturbations. Notably, under this "blurred" versus "aliased" semantic prior, the reconstruction exhibits a different evolution compared with the case presented in **Figure 7**. As shown in **Figure 8(B)**, the reconstruction begins with an aliased appearance due to undersampling and progressively reduces aliasing artifacts throughout the iterations. At later iterations, the reconstruction becomes increasingly smoothed, eventually resembling the TV-regularized compressed sensing result that was used to generate the auxiliary blurred examples. The large language model's responses follow the same progression, transitioning from "*Aliased*" in the early iterations to "*Blurred*" as the reconstruction converges.

**Supporting Information Figure S2** presents reconstruction results obtained under four representative language instructions that are semantically equivalent but phrased differently. The reconstructed images exhibit minimal visual variation across instructions. Quantitatively, the pixel-wise standard deviation across all 20 reconstructions is shown in the first row, with the average remaining below 0.02, corresponding to subtle fluctuations in normalized MRI intensities in the 0-1 range.

## 4  DISCUSSION

In this study, we investigated the potential of vision-language foundation models to enhance MRI reconstruction. We employed the pre-trained Janus model, which provides high-level structured perceptual and conceptual representations by mapping images and languages into a semantic space. Leveraging this capability, we introduced a contrastive objective that guides the reconstruction toward the semantic prior distributions. Two forms of auxiliary information, image-only and image-language, were explored to construct the prior distributions. In the image-only setting, the semantic priors produced reconstructions that better preserved fine anatomical structures and perceptual quality, with good generalizability and robustness. The image-language setting further extended the prior distributions with limited image examples. Additionally, the results demonstrated that the reconstruction can be guided towards image attributes explicitly described in natural language. Together, these findings highlight the promise of vision-language semantic priors in expanding the range of information that can be leveraged for MRI reconstruction, enabling the incorporation of perceptual and linguistic knowledge as a complement to traditional physics-based and mathematically defined constraints.

## 4.1    Results Analysis

The experimental findings offer several insights into how the proposed semantic priors influence MRI reconstruction. In the image-only setting, the semantic priors yield improved visual fidelity (**Figures 2 and 3**) and superior Tenengrad and LPIPS values (**Table 1**), even though PSNR and SSIM are lower. Because PSNR and SSIM primarily emphasize pixel-wise intensity agreement and may not reliably reflect perceived image quality as noted in prior studies,[64] we conducted a reader study to obtain an evaluation from a clinical perspective. The results confirmed that radiologists rated the semantic-prior reconstructions as qualitatively superior across all evaluation criteria. These observations indicate that semantic priors derived from vision-language foundation models encode perceptually meaningful features that complement conventional regularizations. Beyond matched acquisition settings, the proposed priors also exhibit robustness across contrast variations. As shown in **Figure 6**, semantic priors extracted from fat-suppressed images remain effective for reconstructing non-fat-suppressed data, likely because the semantic features capture contextual information that supports more realistic reconstructions. This ability to generalize across contrasts further highlights the flexibility of semantic representations. In the image-language setting, the embedding distributions in **Figures 7(A) and 8(A)** reveal two properties. First, incorporating language instructions with Gaussian perturbations results in an expanded, more expressive semantic embedding space, even with only a small number of image examples (from three subjects). This is because language introduces a high-level axis of variation that cannot be captured by image content alone. Second, the resulting embedding distribution is jointly determined by the images and the language instruction, where the images provide visual content and the language instruction specifies which aspect of the image should be emphasized when constructing the semantic embedding, for example, image quality in our study. The reconstruction process shown in **Figures 7(B) and 8(B)** further demonstrates the effectiveness of these priors. The reconstructed images evolve in a manner consistent with the resulting language descriptions. These results highlight the capacity of image-language information to modulate the reconstruction trajectory and influence the resulting image attributes. The reconstruction consistency across differently worded but semantically identical language instructions (**Supporting Information Figure S2**) highlights the flexibility of language-based priors, suggesting that linguistic information may further extend the semantic distribution and enrich the prior knowledge available for MRI reconstruction.

Taken together, these findings show that image-only and image-language priors offer complementary benefits. To facilitate a direct comparison, **Table 2** summarizes their relative accessibility and reconstruction accuracy. As shown, accessibility increases from same-contrast image priors to cross-contrast priors, and further to image-language priors; however, this increased accessibility is accompanied by a gradual reduction of precision in guiding MRI reconstruction. This trade-off suggests that the optimal prior choice may depend on the availability of auxiliary information and the required reconstruction accuracy for a given task.

### 4.2    Technique Considerations

The proposed auxiliary information-guided reconstruction shares similarities with previous studies that incorporate side information, such as multi-contrast[65–67] or longitudinal MRI scans[47] from the same subject, to enhance fast MRI reconstruction. These approaches provide anatomically aligned and subject-specific priors, but the required side information is not always accessible. In contrast, the semantic priors from foundation models allow for a broader set of auxiliary information, including images from different subjects, different contrasts, and even language descriptions. This expanded accessibility offers greater flexibility but also suffers from reduced precision due to the highly abstract nature of the semantic space. Such imprecision may be further amplified by: (1) the limited auxiliary examples, which may not adequately span the semantic manifold, and (2) the fact that the current foundation model used is trained primarily on natural images with their descriptive captions and lack MRI-specific or quality-assessment-related optimization, leading to biased or suboptimal feature distributions for the MRI reconstruction problem. These challenges could be mitigated by expanding the semantic space using large-scale simulated data generated through Bloch-equation-based modeling that incorporates realistic tissue properties, thereby providing more representative and diverse embeddings. More importantly, addressing these limitations will require a deeper understanding of the semantic representations learned by foundation models, including how to adapt them across domains, enrich and regularize the semantic manifold, and manipulate semantic distributions that align more closely with human perceptual and conceptual frameworks. Recent advances in diffusion models highlight that appropriately structuring and refining latent representations can lead to improved generative performance and controllability.[68–71] Inspired by these developments, exploring similar semantic alignment and feature-space manipulation strategies in vision-language foundation models may not only benefit medical imaging applications but also contribute to the broader development of more robust, interpretable, and intelligent foundation models.

### 4.3 Limitations

Our method has several limitations. First, as discussed above, the foundation model used in this study was not trained on medical images, which may limit its ability to capture the domain-specific semantic representations needed for accurate MRI reconstruction. Future work could address this limitation by fine-tuning such models on large-scale medical image-text datasets or by developing foundation models tailored specifically to medical imaging. Second, the language used to guide reconstruction in this work is limited to simple semantic directions and operates indirectly through contrastive embedding alignment. A promising direction for future research is to explore image-generation foundation models, in which language instructions directly shape the generated image and could therefore serve as explicit constraints within the MRI optimization process. Third, the current framework requires substantial GPU memory. Although the foundation model remains frozen during reconstruction, its large-scale architecture still imposes significant computational demands, potentially limiting scalability and clinical deployment. Techniques such as model compression,[72,73] pruning,[74,75] or distilling the semantic guidance into lightweight networks[76] may help reduce computational cost while retaining performance.

## 5 CONCLUSION

In summary, this work provides a proof-of-concept for incorporating visual-language foundation model priors into MRI reconstruction. The proposed framework demonstrates that semantic embeddings from both images and language can guide the reconstruction toward a perceptually faithful result. By extending the prior space beyond traditional handcrafted features, this approach opens new avenues for instruction-driven, concept-aware reconstruction, laying the groundwork for future research on multimodal guidance in medical imaging.

# 6 ACKNOWLEDGEMENTS

The research reported in this publication was supported by the National Institute of Biomedical Imaging and Bioengineering under Award Number R21EB031185, the National Institute of Arthritis and Musculoskeletal and Skin Diseases under Award Numbers R01AR081344, R01AR079442, and R56AR081017.

# 8 Table

Table 1. Comparison of PSNR, SSIM, LPIPS, and Tenengrad metrics on knee and brain datasets at different acceleration rates. Values are reported as mean±standard deviation, with the best performance in each setting highlighted in bold. Semantic priors achieve better performance in terms of LPIPS and Tenengrad across all architectures.

| | Knee (R=4) | | | | | |
|---|---|---|---|---|---|---|
| Methods | End-to-end | | Unroll | | INR | |
| | TV | Semantic | TV | Semantic | TV | Semantic |
| PSNR (↑) | **30.31±2.33** | 29.47±2.05 | **31.10±2.37** | 30.52±2.26 | 31.27±2.31 | **31.76±2.36** |
| SSIM (↑) | **0.760±0.064** | 0.733±0.063 | **0.757±0.069** | 0.735±0.070 | 0.750±0.074 | **0.768±0.071** |
| LPIPS (↓) | 0.178±0.047 | **0.173±0.040** | 0.155±0.055 | **0.124±0.042** | 0.111±0.038 | **0.096±0.036** |
| Tenengrad (↑) | 0.038±0.014 | **0.043±0.015** | 0.043±0.015 | **0.052±0.018** | 0.049±0.017 | **0.051±0.018** |

| | Knee (R=6) | | | | | |
|---|---|---|---|---|---|---|
| Methods | End-to-end | | Unroll | | INR | |
| | TV | Semantic | TV | Semantic | TV | Semantic |
| PSNR (↑) | **29.39±2.18** | 28.47±2.05 | **30.25±2.21** | 29.42±2.11 | 30.43±2.13 | **31.44±2.30** |
| SSIM (↑) | **0.712±0.069** | 0.679±0.069 | **0.711±0.071** | 0.678±0.072 | 0.726±0.073 | **0.756±0.071** |
| LPIPS (↓) | 0.245±0.048 | **0.224±0.041** | 0.200±0.049 | **0.165±0.042** | 0.155±0.043 | **0.125±0.040** |
| Tenengrad (↑) | 0.030±0.011 | **0.045±0.017** | 0.033±0.011 | **0.050±0.017** | 0.043±0.015 | **0.048±0.017** |

| | Brain (R=4) | | | | | |
|---|---|---|---|---|---|---|
| Methods | End-to-end | | Unroll | | INR | |
| | TV | Semantic | TV | Semantic | TV | Semantic |
| PSNR (↑) | **33.76±2.61** | 32.83±2.32 | **34.34±2.79** | 33.97±2.56 | **34.25±2.70** | 34.09±2.56 |
| SSIM (↑) | **0.885±0.037** | 0.861±0.040 | **0.890±0.044** | **0.890±0.044** | 0.884±0.041 | **0.897±0.040** |
| LPIPS (↓) | 0.110±0.044 | **0.109±0.040** | 0.096±0.043 | **0.092±0.041** | 0.109±0.043 | **0.089±0.040** |
| Tenengrad (↑) | 0.067±0.030 | **0.071±0.032** | 0.078±0.035 | **0.081±0.037** | 0.074±0.033 | **0.077±0.034** |

| | Brain (R=5) | | | | | |
|---|---|---|---|---|---|---|
| Methods | End-to-end | | Unroll | | INR | |
| | TV | Semantic | TV | Semantic | TV | Semantic |
| PSNR (↑) | **32.74±2.48** | 31.78±2.27 | **32.99±2.73** | 32.38±2.54 | **33.32±2.59** | 33.13±2.47 |
| SSIM (↑) | **0.868±0.040** | 0.841±0.045 | **0.870±0.047** | 0.864±0.046 | 0.869±0.045 | **0.881±0.042** |
| LPIPS (↓) | 0.128±0.045 | **0.122±0.042** | 0.114±0.044 | **0.110±0.042** | 0.122±0.045 | **0.100±0.041** |
| Tenengrad (↑) | 0.064±0.030 | **0.068±0.031** | 0.071±0.032 | **0.075±0.034** | 0.072±0.033 | **0.075±0.034** |

**Table 2**. Summary of the semantic priors evaluated in this study, illustrating the trade-off between accessibility and precision.

| Prior type | Accessibility | Accuracy |
| --- | --- | --- |
| Same-contrast images | Low: Requires multiple high-quality MRI images with the same contrast as the reconstructed image. | High: Provides the most accurate guidance, as the prior matches both anatomy and contrast. |
| Different-contrast images | Medium: More accessible, though it still requires multiple images. | Medium: Moderately accurate, as anatomical consistency is preserved, but different contrast introduces variability. |
| Images and language | High: Requires far fewer image samples, with language descriptions providing an efficient mechanism for expanding the semantic embeddings. | Low: Less accurate, as guidance relies on the semantic alignment of language, which is highly abstract. |

# 9 Figure Captions

**Figure 1.** Overview of the proposed semantic distribution guided MRI reconstruction framework. (A) Architecture of vision-language foundation models used to extract semantic embeddings. (B) Deep learning-based MRI reconstruction networks with integrated data consistency constraints. (C) Semantic space optimization using a contrastive objective. Two types of semantic embedding are investigated, including those derived from image-only exemplars and those incorporating image-language auxiliary information. Both the reconstructed output and the auxiliary information are encoded into the same semantic space, where the contrastive objective aligns the reconstructed representation with the target semantic distribution.

**Figure 2.** Reconstruction results on knee data using conventional TV regularization and the proposed semantic prior. (A) Results at an acceleration rate of R = 4. Both approaches effectively suppress aliasing artifacts, while the semantic prior preserves more realistic bone textures and produces images that are visually closer to the fully sampled reference. (B) Results at R = 6. The meniscus tear (yellow arrow) is more clearly depicted when using semantic priors. The layered cartilage (blue arrow) and surrounding tissue (green arrow) also show improved sharpness and structural definition across all three deep learning methods.

**Figure 3.** Reconstruction results on brain data using conventional TV regularization and the proposed semantic prior. (A) Results at an acceleration rate of R=4. The semantic prior improves visual fidelity and enhances fine anatomical details, such as the boundaries of the globus pallidus (red arrow). (B) Results at R=5. In the lesion region, the surrounding structures are more distinctly delineated in the end-to-end and unrolled reconstructions guided by semantic priors (yellow arrow).

**Figure 4.** Reader study results comparing reconstructions with the conventional and the proposed semantic priors across three deep learning architectures. The semantic priors consistently achieve higher subjective scores in Artifacts, Sharpness of Anatomical Structures, Diagnostic Confidence, and Overall Quality, demonstrating improved perceptual quality. Asterisks indicate statistically significant differences ($p < 0.05$).

**Figure 5.** Visualization of embedding distributions and optimization trajectories in the image semantic space. Each point corresponds to the embedding of a single image slice. (A) Clustering of undersampled, fully sampled, and reconstructed data across three hierarchical semantic levels. Reconstructed embeddings (red) lie closer to the fully sampled cluster (orange), indicating successful semantic alignment. (B) Optimization trajectories of a representative example, shown

with color gradients from blue (initial) to red (final). The reconstructed embeddings converge toward the fully sampled cluster, demonstrating effective semantic-space optimization.

**Figure 6.** Reconstruction results on knee data without fat suppression. Reconstruction guided by the conventional regularization and the proposed prior derived from auxiliary fat-suppressed images is compared. The meniscus tear (yellow arrow) and surrounding tissue (red arrow) are more clearly delineated when using the proposed semantic prior.

**Figure 7.** High-quality versus low-quality scenario under the proposed image-language prior for semantic distribution-guided MRI reconstruction at R=4. (A) Language semantic embedding space showing separable high-quality and low-quality clusters, with Gaussian-perturbed embeddings extending the semantic distribution. A representative reconstruction trajectory transitions from the low-quality to the high-quality region. (B) Display of the reconstruction evolution at different iterations. The reconstructed images progressively improve, accompanied by a shift in the language model's assessments from "Low-quality" to "High-quality."

**Figure 8**. Further validation using the blurred versus aliased case under the proposed image-language prior for MRI reconstruction at R=4. (A) Language semantic embedding space showing separable blurred and aliased clusters, with Gaussian-perturbed embeddings extending the semantic distribution. A representative reconstruction trajectory transitions from the aliased to the blurred region. (B) Display of the reconstruction evolution at different iterations. The reconstructed image progressively reduces aliasing artifacts as it becomes increasingly smoothed, and the language model's assessments shift accordingly from "Aliased" to "Blurred."

**Supporting Information Figure S1**. Reconstruction results at 4× acceleration on cases with substantial anatomical abnormalities. Across all three deep-learning methods, semantic priors enable recovery of major pathological features, whereas the zero-filled images show severe aliasing and loss of anatomical detail.

**Supporting Information Table S1.** Twenty linguistically varied but semantically equivalent language instructions used to construct the embedding distribution. Each instruction prompts a binary judgment of whether an image is high-quality or low-quality, providing diverse textual formulations that express the same underlying semantic intent for contrastive objective.

**Supporting Information Figure S2.** Reconstruction results under four representative language instructions that are semantically equivalent but phrased differently. The reconstructed images show minimal visual variation across instructions. The pixel-wise standard deviation is calculated across all 20 reconstructions, indicating subtle fluctuations across these instructions.

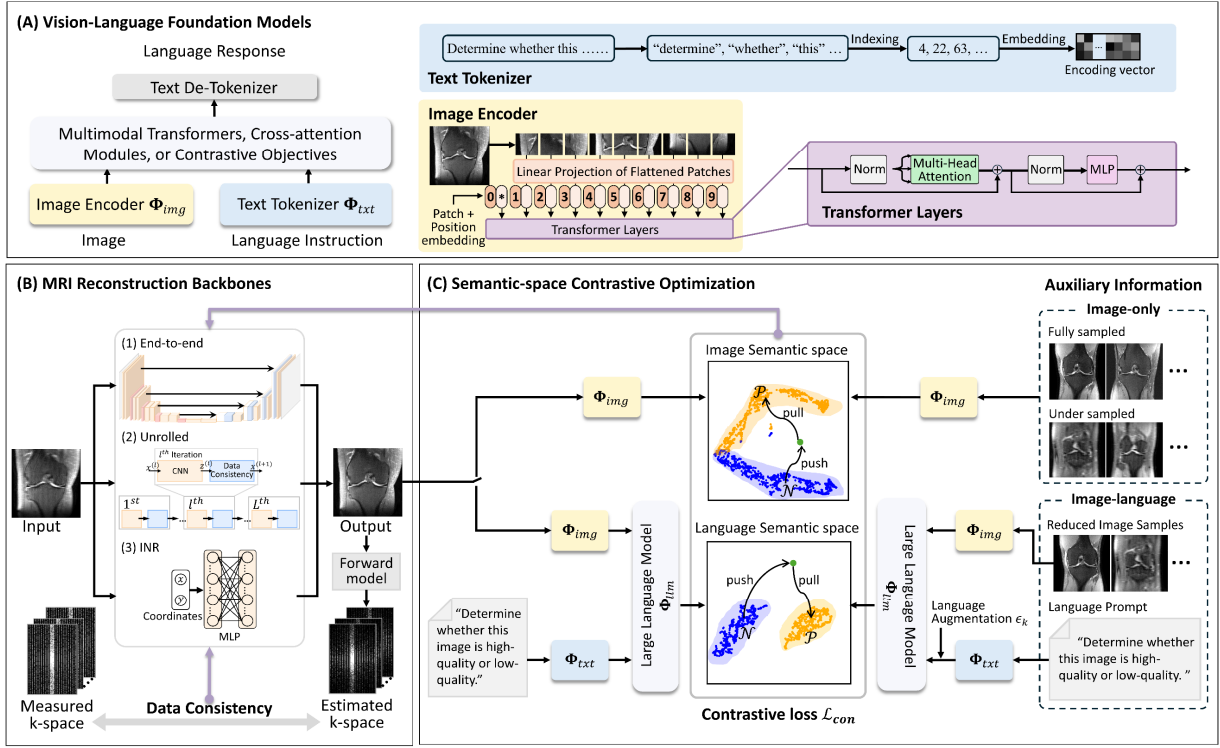

**Figure 1.** Overview of the proposed semantic distribution guided MRI reconstruction framework. (A) Architecture of vision-language foundation models used to extract semantic embeddings. (B) Deep learning-based MRI reconstruction networks with integrated data consistency constraints. (C) Semantic space optimization using a contrastive objective. Two types of semantic embedding are investigated, including those derived from image-only exemplars and those incorporating image-language auxiliary information. Both the reconstructed output and the auxiliary information are encoded into the same semantic space, where the contrastive objective aligns the reconstructed representation with the target semantic distribution.

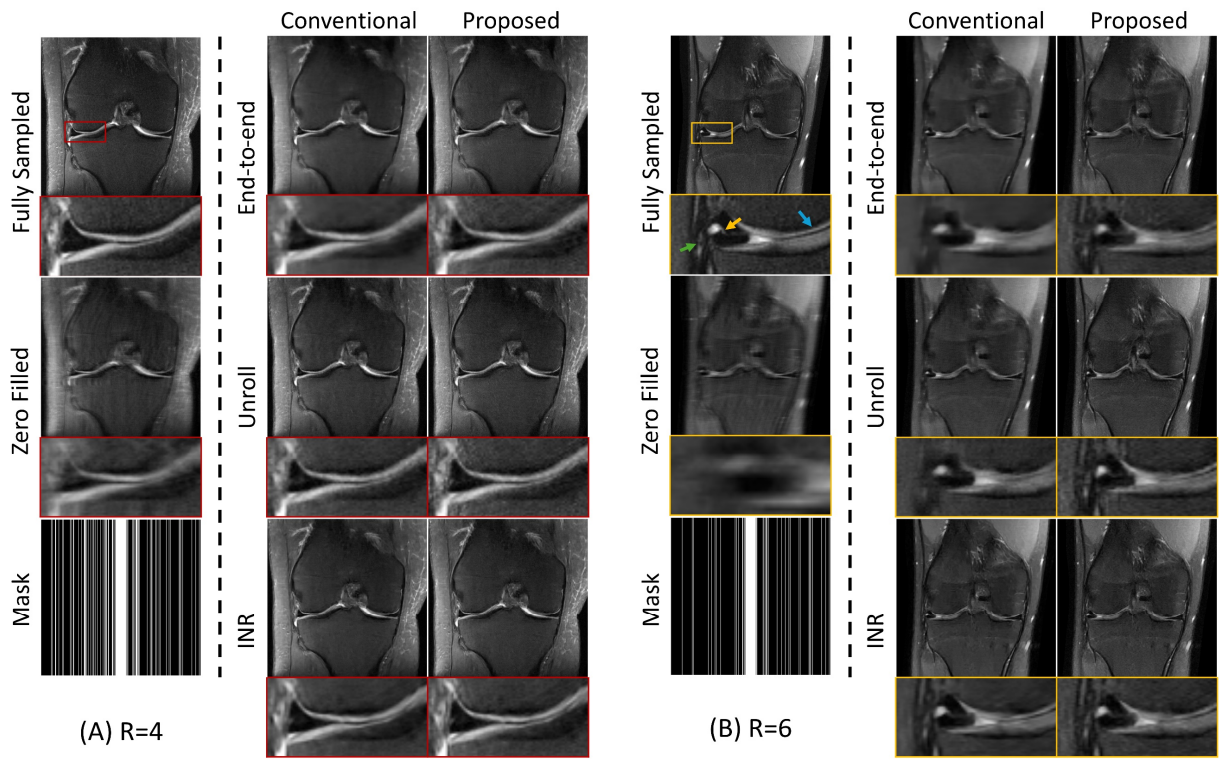

**Figure 2**. Reconstruction results on knee data using conventional TV regularization and the proposed semantic prior. (A) Results at an acceleration rate of R = 4. Both approaches effectively suppress aliasing artifacts, while the semantic prior preserves more realistic bone textures and produces images that are visually closer to the fully sampled reference. (B) Results at R = 6. The meniscus tear (yellow arrow) is more clearly depicted when using semantic priors. The layered cartilage (blue arrow) and surrounding tissue (green arrow) also show improved sharpness and structural definition across all three deep learning methods.

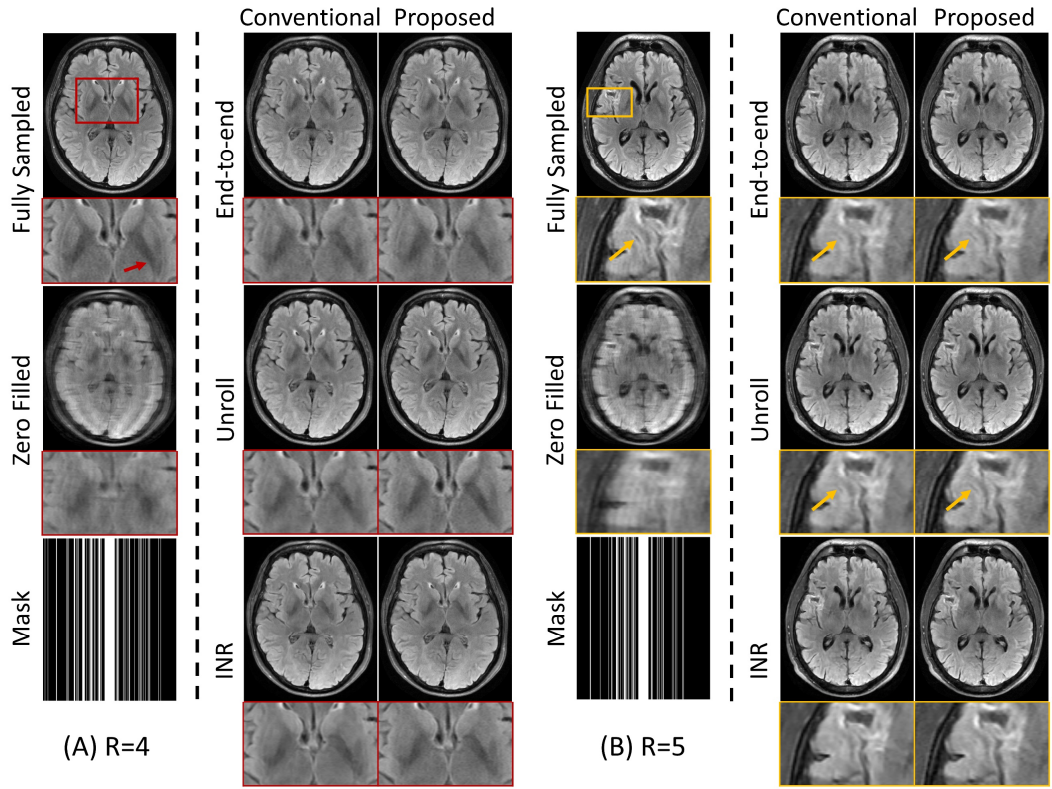

**Figure 3**. Reconstruction results on brain data using conventional TV regularization and the proposed semantic prior. (A) Results at an acceleration rate of R=4. The semantic prior improves visual fidelity and enhances fine anatomical details, such as the boundaries of the globus pallidus (red arrow). (B) Results at R=5. In the lesion region, the surrounding structures are more distinctly delineated in the end-to-end and unrolled reconstructions guided by semantic priors (yellow arrow).

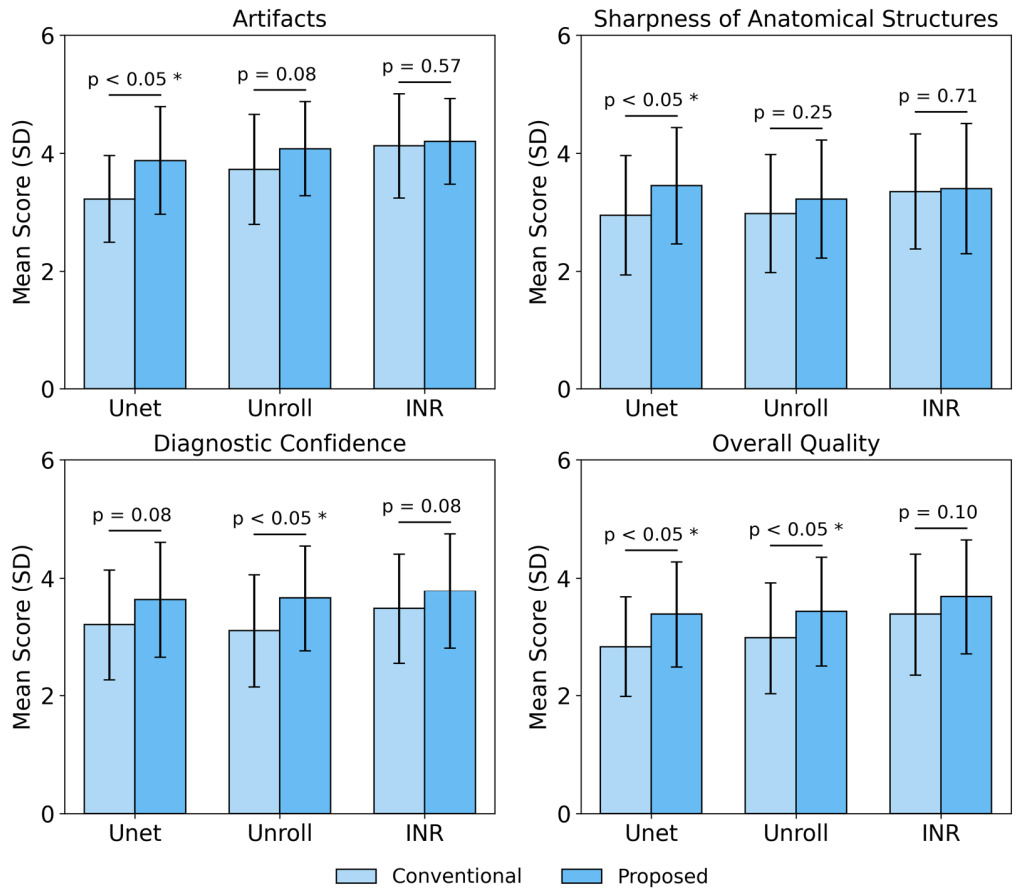

**Figure 4.** Reader study results comparing reconstructions with the conventional and the proposed semantic priors across three deep learning architectures. The semantic priors consistently achieve higher subjective scores in Artifacts, Sharpness of Anatomical Structures, Diagnostic Confidence, and Overall Quality, demonstrating improved perceptual quality. Asterisks indicate statistically significant differences (p < 0.05).

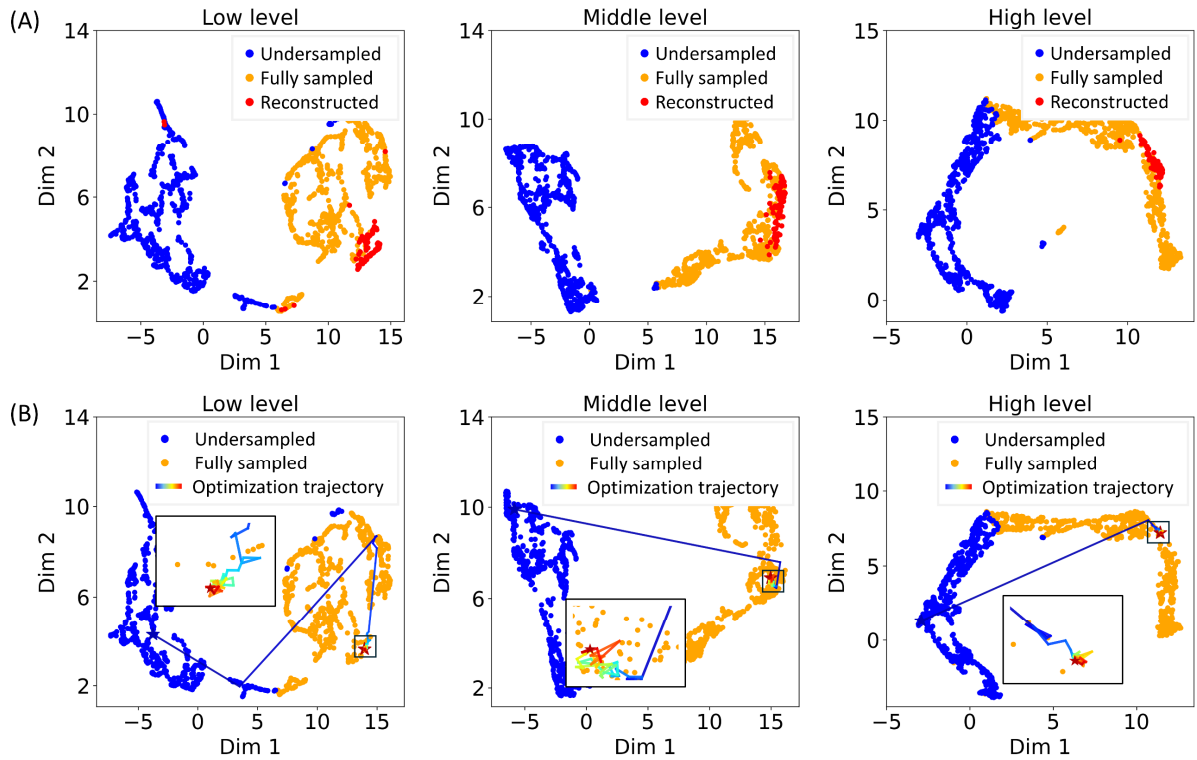

**Figure 5.** Visualization of embedding distributions and optimization trajectories in the image semantic space. Each point corresponds to the embedding of a single image slice. (A) Clustering of undersampled, fully sampled, and reconstructed data across three hierarchical semantic levels. Reconstructed embeddings (red) lie closer to the fully sampled cluster (orange), indicating successful semantic alignment. (B) Optimization trajectories of a representative example, shown with color gradients from blue (initial) to red (final). The reconstructed embeddings converge toward the fully sampled cluster, demonstrating effective semantic-space optimization.

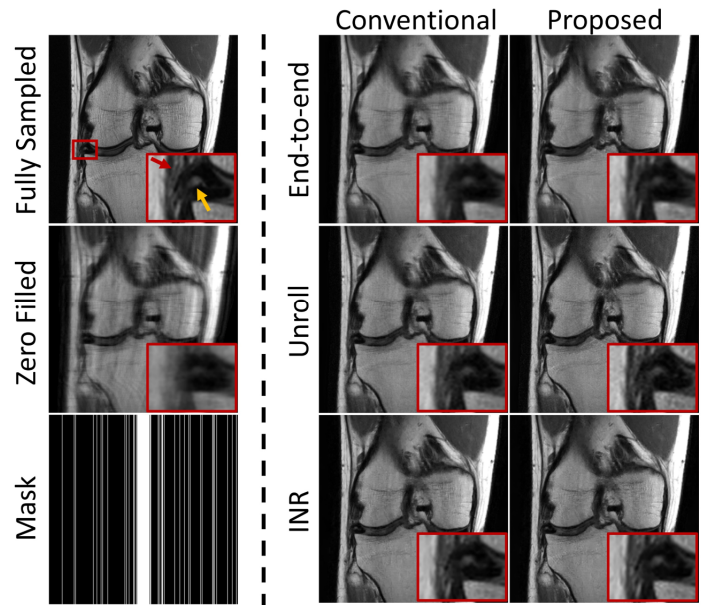

**Figure 6.** Reconstruction results on knee data without fat suppression. Reconstruction guided by the conventional regularization and the proposed prior derived from auxiliary fat-suppressed images is compared. The meniscus tear (yellow arrow) and surrounding tissue (red arrow) are more clearly delineated when using the proposed semantic prior.

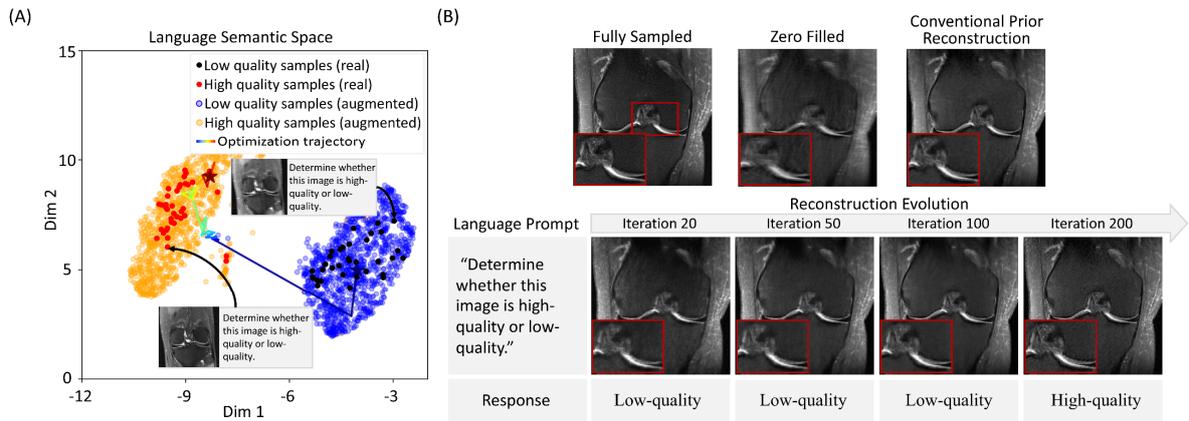

**Figure 7.** High-quality versus low-quality scenario under the proposed image-language prior for semantic distribution-guided MRI reconstruction at R=4. (A) Language semantic embedding space showing separable high-quality and low-quality clusters, with Gaussian-perturbed embeddings extending the semantic distribution. A representative reconstruction trajectory transitions from the low-quality to the high-quality region. (B) Display of the reconstruction evolution at different iterations. The reconstructed images progressively improve, accompanied by a shift in the language model's assessments from "Low-quality" to "High-quality."

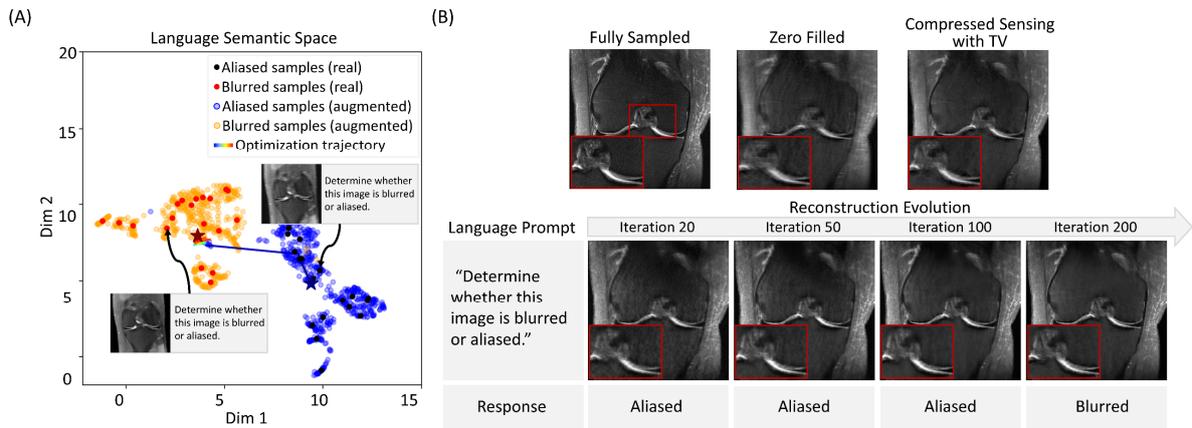

**Figure 8**. Further validation using the blurred versus aliased case under the proposed image-language prior for MRI reconstruction at R=4. (A) Language semantic embedding space showing separable blurred and aliased clusters, with Gaussian-perturbed embeddings extending the semantic distribution. A representative reconstruction trajectory transitions from the aliased to the blurred region. (B) Display of the reconstruction evolution at different iterations. The reconstructed image progressively reduces aliasing artifacts as it becomes increasingly smoothed, and the language model's assessments shift accordingly from "Aliased" to "Blurred."

# Supporting Information

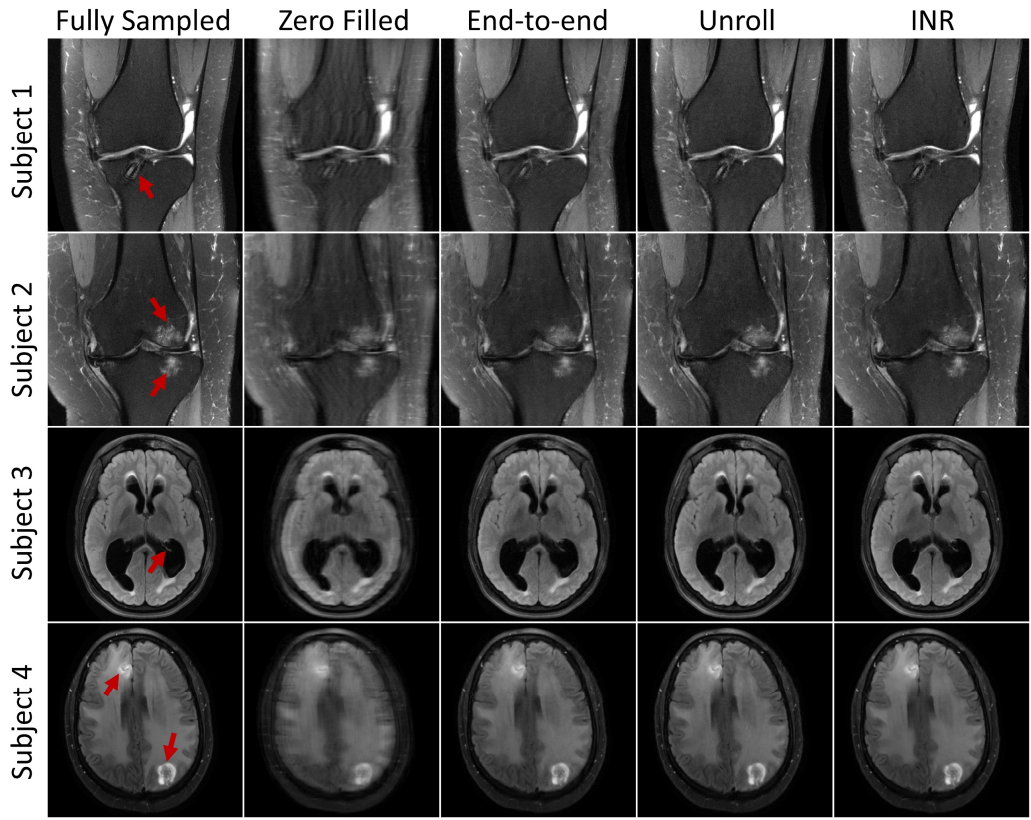

**Supporting Information Figure S1**. Reconstruction results at 4× acceleration on cases with substantial anatomical abnormalities. Across all three deep-learning methods, semantic priors enable recovery of major pathological features, whereas the zero-filled images show severe aliasing and loss of anatomical detail.

**Supporting Information Table S1.** Twenty linguistically varied but semantically equivalent language instructions used to construct the embedding distribution. Each instruction prompts a binary judgment of whether an image is high-quality or low-quality, providing diverse textual formulations that express the same underlying semantic intent for contrastive objective.

| ID | Language Instructions |
|---|---|
| 1 | Determine whether this image is high-quality or low-quality. |
| 2 | Decide if the image is best described as high-quality or low-quality. |
| 3 | Does this image appear high or low quality? |
| 4 | Assess whether this image is high quality or low quality. |
| 5 | Decide if the image appears high quality or low quality. |
| 6 | Evaluate whether the image is high-quality or low-quality. |
| 7 | Indicate if this image represents high quality or low quality. |
| 8 | Choose whether this image fits high quality or low quality. |
| 9 | Check whether this image is high-quality or low-quality. |
| 10 | Is the image high or low quality? |
| 11 | Provide a classification of the image quality: high or low. |
| 12 | Give a classification of the image quality: high or low. |
| 13 | Does this look like a high-quality or low-quality image? |
| 14 | Would you judge the image as high-quality or low-quality? |
| 15 | Is the quality of this image high-quality or low-quality? |
| 16 | Decide whether the image looks high-quality or low-quality. |
| 17 | Assess whether the overall appearance is high-quality or low-quality. |
| 18 | Indicate whether this image is high-quality or low-quality. |
| 19 | Decide if the image should be judged high-quality or low-quality. |
| 20 | Classify the image quality: high or low. |

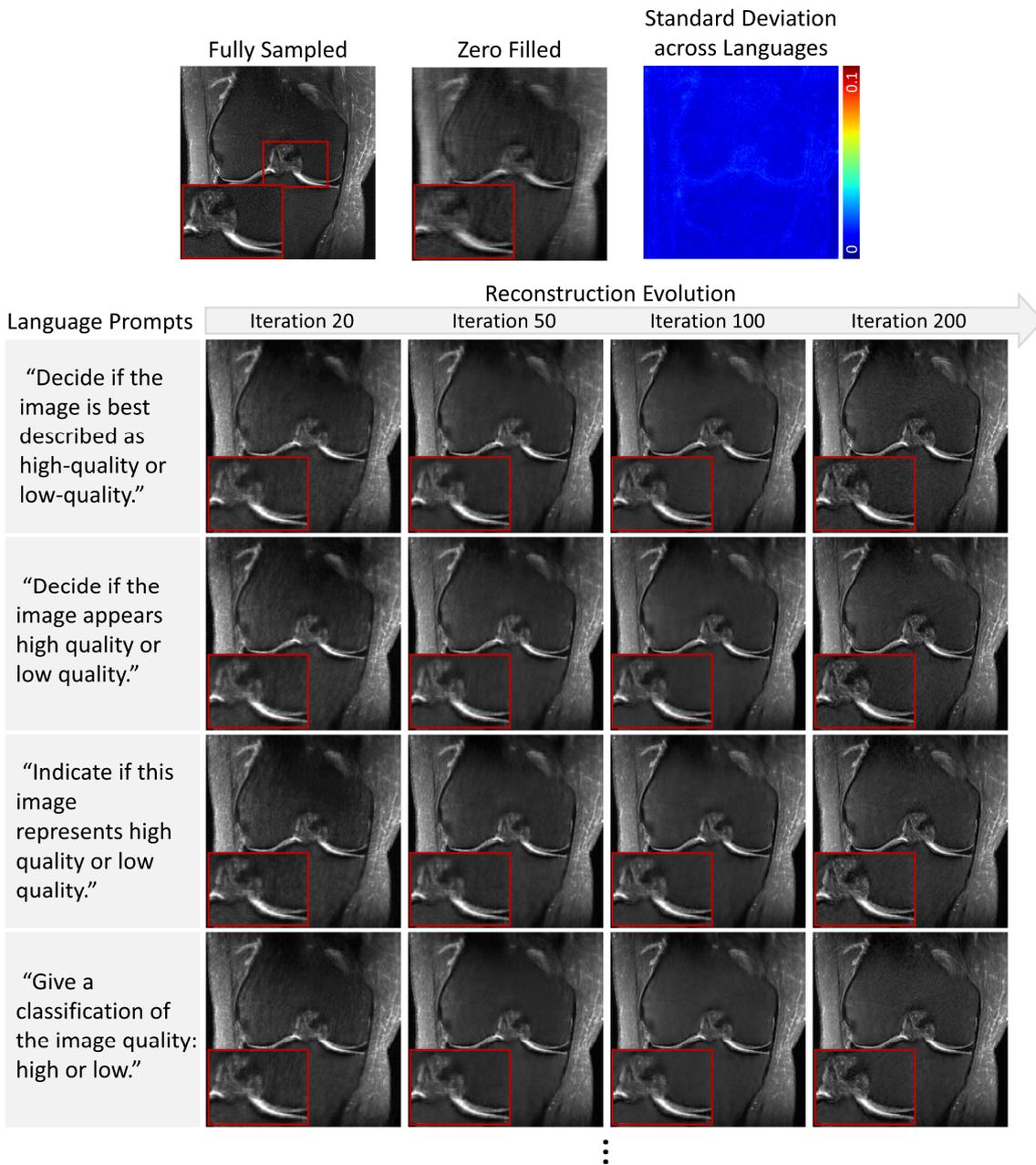

**Supporting Information Figure S2.** Reconstruction results under four representative language instructions that are semantically equivalent but phrased differently. The reconstructed images show minimal visual variation across instructions. The pixel-wise standard deviation is calculated across all 20 reconstructions, indicating subtle fluctuations across these instructions.